\documentclass[conference]{IEEEtran}
\IEEEoverridecommandlockouts
\usepackage{cite}
\usepackage{amsmath,amssymb,amsfonts}
\usepackage{algorithmic}
\usepackage{graphicx}
\usepackage{textcomp}
\usepackage{xcolor}
\def\BibTeX{{\rm B\kern-.05em{\sc i\kern-.025em b}\kern-.08em
    T\kern-.1667em\lower.7ex\hbox{E}\kern-.125emX}}
\begin{document}

\title{Resource-Efficient Medical Report Generation using Large Language Models
}

\author{\IEEEauthorblockN{Abdullah, Ameer Hamza, Seong Tae Kim}
\IEEEauthorblockA{\textit{Department of Computer Science and Engineering} \\
\textit{Kyung Hee University}\\
Yongin, Republic of Korea \\
abdullahijaz@khu.ac.kr, ameer@khu.ac.kr, st.kim@khu.ac.kr}
}

\maketitle

\begin{abstract}
Medical report generation is the task of automatically writing radiology reports for chest X-ray images. Manually composing these reports is a time-consuming process that is also prone to human errors. Generating medical reports can therefore help reduce the burden on radiologists. In other words, we can promote greater clinical automation in the medical domain. In this work, we propose a new framework leveraging vision-enabled Large Language Models (LLM) for the task of medical report generation. We introduce a lightweight solution that achieves better or comparative performance as compared to previous solutions on the task of medical report generation. We conduct extensive experiments exploring different model sizes and enhancement approaches, such as prefix tuning to improve the text generation abilities of the LLMs. We evaluate our approach on a prominent large-scale radiology report dataset - MIMIC-CXR. Our results demonstrate the capability of our resource-efficient framework to generate patient-specific reports with strong medical contextual understanding and high precision. 
\end{abstract}


\section{Introduction}
Medical report generation aims to automatically generate detailed paragraphs that describe the observations and findings from a given chest X-ray image. Writing these radiology reports manually is a time-consuming task that is also prone to errors. Automating the medical report generation process can relieve radiologists of this workload and promote clinical automation.

In recent years, researchers have proposed various solutions to the task of medical report generation. These solutions can be broadly categorized into two main approaches. The first approach focuses on improving the model structure to enhance the performance of medical report generation. For instance, some works have utilized hierarchically structured LSTM \cite{b1} architectures to handle the long-form nature of medical reports. Others have explored different network structures, such as using a generative sentence model and a generative paragraph model that leverages the generated sentences to produce the next sentence \cite{b2}. Additionally, image-report matching networks have been proposed \cite{b3} to bridge the gap between the image and the text. More recently, transformer \cite{b4} architectures have been used as decoders, along with memory mechanisms, as an alternative to LSTM-based models \cite{b5}. Some approaches have also leveraged LLMs \cite{b6} to harness their generative capabilities for medical report generation. However, these works incorporated very large LLMs and also trained those LLMs which is resource-intensive and this hinders their adoption in clinical automation. In comparison, we provide a lightweight solution which achieves better or similar performance on the natural language generation metrics while being resource-efficient.

The second category of solutions focuses on leveraging available medical domain knowledge to improve the quality of the generated reports. This includes integrating knowledge graphs \cite{b7}, utilizing disease tags \cite{b8}, and combining general knowledge from pre-constructed knowledge graphs with specific knowledge derived from retrieving similar reports \cite{b9}. Some works have also proposed using a medical concepts generation network \cite{b10} to produce semantic information and integrate it into the report generation process.

In our work as shown in Fig. 1, we aim to promote clinical automation by introducing a resource-efficient framework which incorporates LLM and enhances its capabilities without fine-tuning of LLM by utilizing prefix tuning for the task of medical report generation.

\section{Methods}

The proposed method largely consists of a vision encoder, a large language model, and a mapping network. For efficient training of the whole model, the mapping network is the only trainable part of our model. The details will be introduced in the following subsections. 
\subsection{Vision Encoder}
Contrastive language image pretraining (CLIP) \cite{b11} introduces a method for training a text and image-based encoder model on multimodal data using contrastive learning. This approach aims to bring similar data pairs closer together in the model's projection space while pushing dissimilar pairs farther apart, thereby bridging the gap between different modalities such as text and image. As a result, closely related text and image samples exhibit high cosine similarity scores, while dissimilar pairs show lower scores.

We utilize a medical CLIP (i.e., MedCLIP \cite{b12}), to extract visual embeddings, which we term prefix embeddings for use in the medical report generation. MedCLIP is trained to capture detailed information present in chest X-ray images. When these visual embeddings are translated through a lightweight mapping network into the language model's space, they provide valuable visual context. This enables the model to generate patient-specific radiology reports.\par

\begin{figure}
    \centering
    \includegraphics[width=1\linewidth]{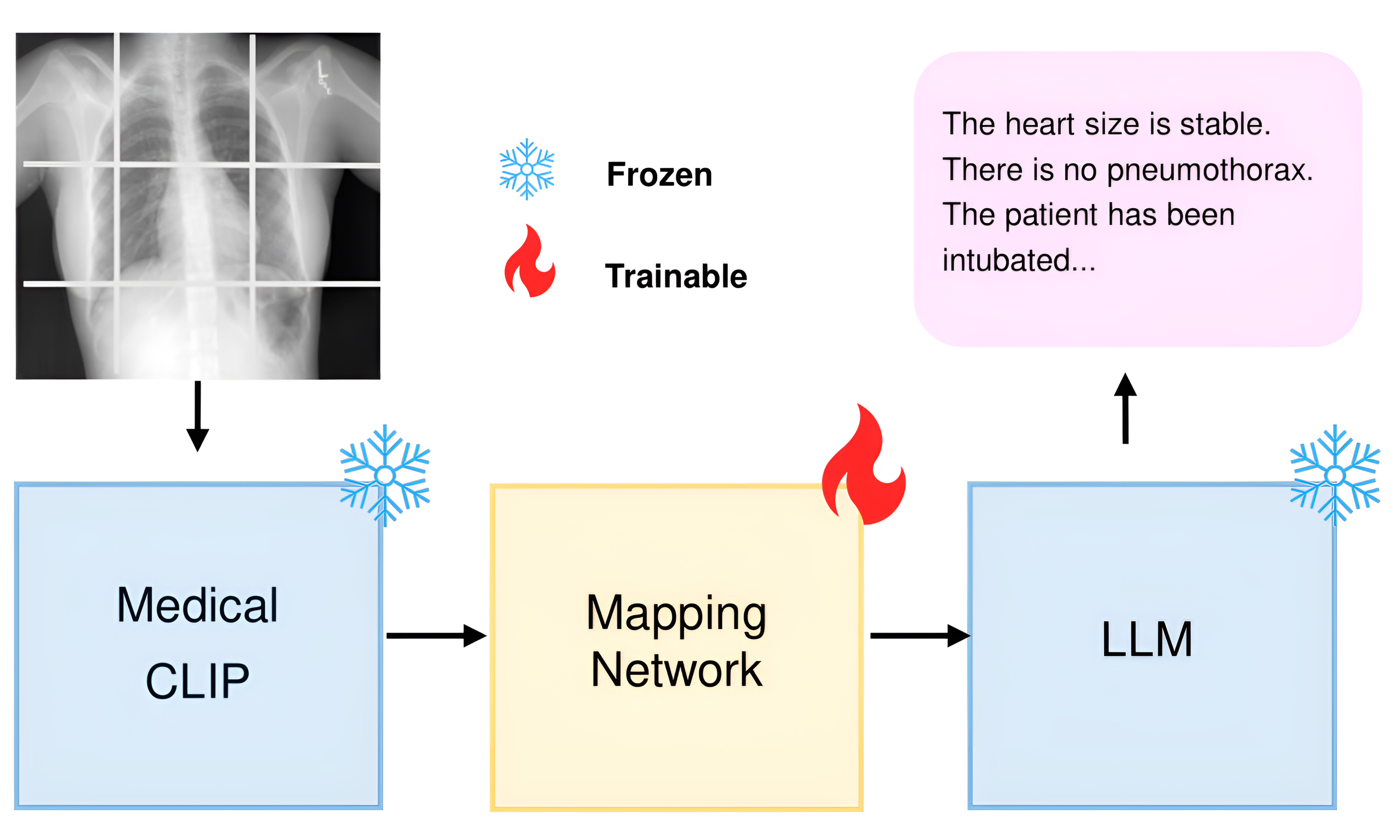}
    \caption{Figure of our proposed framework consisting of a vision encoder, a large language model, and a mapping network.}
    \label{fig:enter-label}
\end{figure}

\subsection{Large Language Models}
Large language models are a class of powerful neural networks that are trained on vast amounts of text data, enabling them to generate human-like text and perform a wide range of natural language processing tasks. To evaluate the performance of our method, we utilized a variety of LLM models. Specifically, we employed the pioneering GPT-2 \cite{b13} models as well as the more recent Qwen1.5 \cite{b14} model.
Both GPT-2 and Qwen1.5 are LLMs, but they differ in their training datasets. The GPT-2 models were trained on the WebText dataset, which comprises 8 million web pages. The training objective for these models was next-word prediction, which is a common objective for autoregressive models. In contrast, Qwen1.5 is an LLM model that has demonstrated better performance on natural language generation tasks, even with a relatively smaller model size.
By using both GPT-2 and Qwen1.5 LLMs in our experiments, we aimed to explore the performance of our method on different LLMs and understand how the choice of pre-trained LLM can impact the overall results for medical report generation. The LLMs in our framework are frozen and thus reduce the requirement of resources making them resource-efficient.

\subsection{Mapping Network}
Prefix Tuning, as proposed by \cite{b15}, involves adding task-specific vectors at the beginning of the input sequence of language models while keeping the model parameters fixed. This only optimizes the prefix while the LLM is frozen. The language model is conditioned on this prefix, along with other information, to generate the medical report. Since we are integrating the language model with vision, we also employ a small mapping network. This network's role is to translate vision features from the medical domain-adapted vision model to the language space of the GPT-based model.

The mapping network is the only trainable part of our framework.
Drawing inspiration from the \cite{b16} approach, we employ a compact transformer-based architecture to convert visual information from CLIP's embedding space to the LLM embedding space. This Mapping Network has a small number of parameters and is trained to effectively translate from CLIP's 512 dimensions to LLM's 768 dimensions for GPT-2 based models and 1024 dimensions for the Qwen1.5 model. By integrating this mapping network, the LLM gains visual capabilities, enhancing its performance in generating reports from chest X-ray images. The visual embeddings from MedCLIP are converted into sequences of tokens based on the hyperparameter of clip length. A trainable prefix is also added before these tokens and these are then processed by the transformer.  The prefix retrieves meaningful information from CLIP embedding through the multi-head attention and it learns to adjust the fixed LLM to the new data.

\section{Experimental Results}
\subsection{Dataset}

For our work, we utilized the publicly available dataset: MIMIC-CXR \cite{b17}. The MIMIC-CXR dataset contains 377,110 chest X-ray images and their corresponding free-text reports. We adopted the standard dataset split of MIMIC-CXR for our experiments. As a preprocessing step, we removed any special characters from the reports and converted all tokens to lowercase. The datasets include a variety of image views, such as frontal and lateral views. However, given the dominance of the anteroposterior (AP) and posteroanterior (PA) views, we focused our experiments solely on the unique AP and PA view image and report pairs which amounted to 243,334. The number of samples for train, val, and test sets is 237,972, 1,959, and 3,403 respectively.

\subsection{Results}
Our method showed improved performance on the task of medical report generation. The GPT-2 based LLMs performed relatively lower while the Qwen1.5 LLM performed better showing improved performance on the NLG metrics \cite{b18} such as Bleu \cite{b19} scores as shown in TABLE \ref{tab:table1}.\par

\begin{table}
\caption{NLG scores for our framework with different LLMs on the task of medical report generation. Params denotes the number of parameters required to make inference.
}\label{tab:table1}
    \centering
    \begin{tabular}{clcccc}
         \hline
  LLM& Params.&Bleu-1& Bleu-2& Bleu-3& Bleu-4\\
\hline
  GPT-2 small& 225M&23.1& 12.4& 7.3& 5.0\\
  GPT-2 medium& 456M&24.6& 13.3& 7.8& 4.9\\
  Qwen1.5-0.5B& 601M&\textbf{34.2}& \textbf{20.0}& \textbf{13.0}& \textbf{8.5}\\
\hline
    \end{tabular}
    
\end{table}

Despite the effectiveness of GPT-2 LLMs in various language tasks, including text generation, their performance on specific NLG metrics such as Bleu scores was not as strong as expected. This could be attributed to their architecture, which might not be optimized for certain types of NLG tasks, such as report generation from medical images.\par

In contrast, the Qwen1.5 LLM exhibited significantly improved performance on the NLG metric compared to the GPT-2 models. The Qwen1.5 model appears to be more effective for tasks requiring generation from multimodal inputs. This improved performance suggests that the Qwen1.5 model can generate more accurate and fluent text, especially in the context of medical report generation from chest X-ray images.\par

\begin{table}
\caption{NLG scores for our Framework with different LLMs on the task of medical report generation.
}\label{tab:table2}
    \centering
    \begin{tabular}{ccccc}
         \hline
  Training&Bleu-1& Bleu-2& Bleu-3& Bleu-4\\
\hline
  Fine Tuning&33.6& 19.9& 12.7& \textbf{8.6}\\
  Prefix Tuning&\textbf{34.2}& \textbf{20.0}& \textbf{13.0}& 8.5\\
\hline
    \end{tabular}
    
\end{table}

TABLE \ref{tab:table2} highlights the effectiveness of Prefix Tuning as compared to Fine Tuning of the LLM. Prefix Tuning showed improved performance and is more resource-efficient as compared to the finetuning of the framework in which the LLM is also trained along with the mapping network. This shows that by leveraging the pretrained LLMs we can get better results on the task of medical report generation. This will enable more clinical automation as the LLMs get efficient in their sizes and become better in their performance in the future.

The Qwen1.5 LLM even outperformed past full transformer-based frameworks and other larger LLM-based solutions on Bleu metric. This is a noteworthy result, as past transformer-based and LLM-based frameworks were resource-intensive and were specifically designed for handling complex NLG tasks. The Qwen1.5 model's ability to surpass these frameworks indicates its effectiveness in integrating visual information and generating high-quality text outputs while being resource-efficient, showing the strength of our framework for the medical report generation task.\par

The results presented in TABLE \ref{tab:table3} demonstrate the potential of our proposed method for medical report generation as compared to full transformer-based and LLM-based frameworks. R2gen(base) [5] denotes the vanilla Transformer, with three layers, 8 heads, and 512 hidden units without other extensions and modifications while RadDialog-INS [6] denotes the Vicuna-7b \cite{b20} based framework trained on the Instruction dataset for medical report generation. Our approach consists of Qwen1.5-0.5B as the LLM. Notably, our approach, which consisted of 601 million total parameters including 101M of trainable parameters of the mapping network as compared to 72 million and an astounding 7 billion parameters of previous studies, was able to achieve competitive performance compared to the more resource-intensive and complex frameworks.\par

At inference time it would require approximately 1GB of VRAM to load the transformer-based models, 2GB to load our Qwen1.5-0.5B LLM-based framework, and a huge number of 25GB VRAM to load the Vicuna-7b LLM-based solutions to generate the medical reports \cite{b21}. This hinders the adoption of these solutions in the medical domain pertaining to their resource requirements. Our framework provides a resource-efficient alternative to these frameworks while achieving better performance.\par

A key strength of our method is its performance on the Bleu-1, Bleu-2, and Bleu-3 metrics, which measure the n-gram overlap between the generated text and reference reports. Our model achieved Bleu-1, Bleu-2, and Bleu-3 scores of 34.2, 20.0, and 13.0 respectively, outperforming the more sophisticated transformer-based and LLM-based models included in the comparison.\par

Our proposed framework represents an efficient approach to medical report generation by leveraging the generative abilities of LLMs efficiently through the use of prefix tuning demonstrating its superior performance on  Natural Language Generation (NLG) metrics specifically the Bleu score compared to full transformer-based and larger LLM-based frameworks.

\begin{table}
\caption{comparison between the number of parameters, VRAM required for inference, and NLG scores of our framework compared to previous works of LLM and Transformer-based Methods on the task of medical report generation. Params denotes the number of parameters required to make inference.
}\label{tab:table3}
    \centering
    \begin{tabular}{l|c|c|c|c|c|c}\hline
 Method&Params.&VRAM&B1& B2& B3& B4\\\hline

R2gen(base) [5]&72M&1GB&31.4& 19.2 & 12.7& 9.0 \\
 RaDialog-INS [6]&7B&25GB&34.0& -& -& \textbf{9.7}\\
Ours&601M&2GB&\textbf{34.2}& \textbf{20.0}& \textbf{13.0}& 8.5\\
\hline

\end{tabular}
    
\end{table}

\section{Conclusion and Future Directions}
Our findings suggest that a well-designed smaller LLM-based framework without using huge resources can capture important linguistic patterns and coherence in the medical report generation task, as compared to full transformer-based and larger LLM-based architectures. This could be particularly advantageous in clinical automation where computational resources are constrained, as our method provides a more resource-efficient solution while maintaining strong performance.
The ability to leverage smaller LLMs, without relying on the full transformer-based structures and larger LLMs, opens up interesting possibilities for further research and optimization. Exploring ways to further enhance our approach, such as through improved pretraining, fine-tuning techniques, or the incorporation of domain-specific knowledge, could lead to even stronger performance on medical report generation.
Additionally, investigating the generalizability of our method to other text generation tasks in the medical domain or beyond would be a valuable direction for future work. Assessing its performance on a wider range of benchmarks could provide further insights into the capabilities and limitations of this more lightweight approach.

Furthermore, we plan to enhance our framework by incorporating more efficient LLMs and training strategies. In the future, we also aim to investigate the use of knowledge graphs as additional knowledge for medical report generation. By continually refining our framework, we aim to provide healthcare professionals with a powerful tool for generating accurate and comprehensive medical reports efficiently.

\section*{Acknowledgement}
This work was supported in part by the Institute of Information and Communications Technology Planning and Evaluation (IITP) Grant funded by the Korea Government (MSIT) under Grant 2022-0-00078 (Explainable Logical Reasoning for Medical Knowledge Generation), Grant RS-2022-00155911 (Artificial Intelligence Convergence Innovation Human Resources Development (Kyung Hee University)), and by the National Research Foundation of Korea(NRF) grant funded by the Korea government(MSIT) (No.RS-2024-00334321).

\end{document}